\pdfoutput=1

\documentclass[11pt]{article}

\usepackage{ACL2023}

\usepackage{times}
\usepackage{latexsym}
\usepackage{amssymb}
\usepackage{amsmath} 
\usepackage{makecell}
\usepackage{tabularx}
\usepackage{booktabs}
\usepackage[inkscapelatex=false]{svg}
\usepackage{subfig}
\usepackage{multicol}
\usepackage{multirow}
\usepackage{graphicx}

\usepackage[T1]{fontenc}

\usepackage[utf8]{inputenc}

\usepackage{microtype}

\usepackage{inconsolata}

\title{ResLoRA: Identity Residual Mapping in Low-Rank Adaption}

\author{
  Shuhua Shi\thanks{~~~Work done during internship at Microsoft.}~~$^{1}$~  
  Shaohan Huang$^{2}$~
  Minghui Song$^{2}$~
  Zhoujun Li\thanks{~~~Corresponding author.}~~$^{1}$~ \\
  \textbf{Zihan Zhang$^{2}$~ Haizhen Huang$^{2}$~ Furu Wei$^{2}$~ Weiwei Deng$^{2}$~ Feng Sun$^{2}$~ Qi Zhang$^{2}$ }\\
  School of Computer Science and Engineering, Beihang University, Beijing, China$^{1}$ \\
  Microsoft$^{2}$ \\
  \texttt{\{shishuhua,lizj\}@buaa.edu.cn} \\
  \texttt{\{shaohanh,minghuisong,zihzha,hhuang,fewer,dedeng,sunfeng,zhang.qi\}@microsoft.com} 
}

\begin{document}
\maketitle
\begin{abstract}
As one of the most popular parameter-efficient fine-tuning (PEFT) methods, low-rank adaptation (LoRA) is commonly applied to fine-tune large language models (LLMs). However, updating the weights of LoRA blocks effectively and expeditiously is challenging due to the long calculation path in the original model. To address this, we propose ResLoRA, an improved framework of LoRA. By adding residual paths during training and using merging approaches to eliminate these extra paths during inference, our method can achieve better results in fewer training steps without any extra trainable parameters or inference cost compared to LoRA. The experiments on NLG, NLU, and text-to-image tasks demonstrate the effectiveness of our method. To the best of our knowledge, ResLoRA is the first work that combines the residual path with LoRA. The code of our method is available at \url{https://github.com/microsoft/LMOps/tree/main/reslora}.
\end{abstract}

\section{Introduction}
%
In recent years, large language models (LLMs) \citep{llm-overview} with hundreds of billions of parameters have shown remarkable performance on various tasks. Fine-tuning LLMs on specific datasets typically leads to better performance than merely giving instructions in the prompt during inference\citep{finetune}. However, the cost of it is often prohibitive due to the large number of parameters involved.

To address this problem, various parameter-efficient fine-tuning (PEFT) methods have been proposed. PEFT methods freeze all parameters in the original model, and only tune a few parameters in the newly added modules. Among them, one of the most popular PEFT methods is LoRA\citep{lora}, which stands for low-rank adaptation. LoRA uses two matrices parallel to the original frozen linear layer with few trainable parameters during training, and merges them together during inference. LoRA incurs no cost in terms of time and computation after merging, and has been mathematically proven \citep{loramath} to be effective, so it has a wide range of applications.

The basic LoRA method still has some limitations. Previous studies mainly focused on either dynamically adjusting the rank of LoRA modules in different layers of the model\citep{adalora}, or using fewer trainable parameters to achieve a similar effect as the original LoRA method\citep{dylora}. However, they overlooked a potential problem: a long backward path hinders the updating of parameters in LoRA blocks.

\begin{figure}[t]
  \centering
  \includegraphics[width=\linewidth]{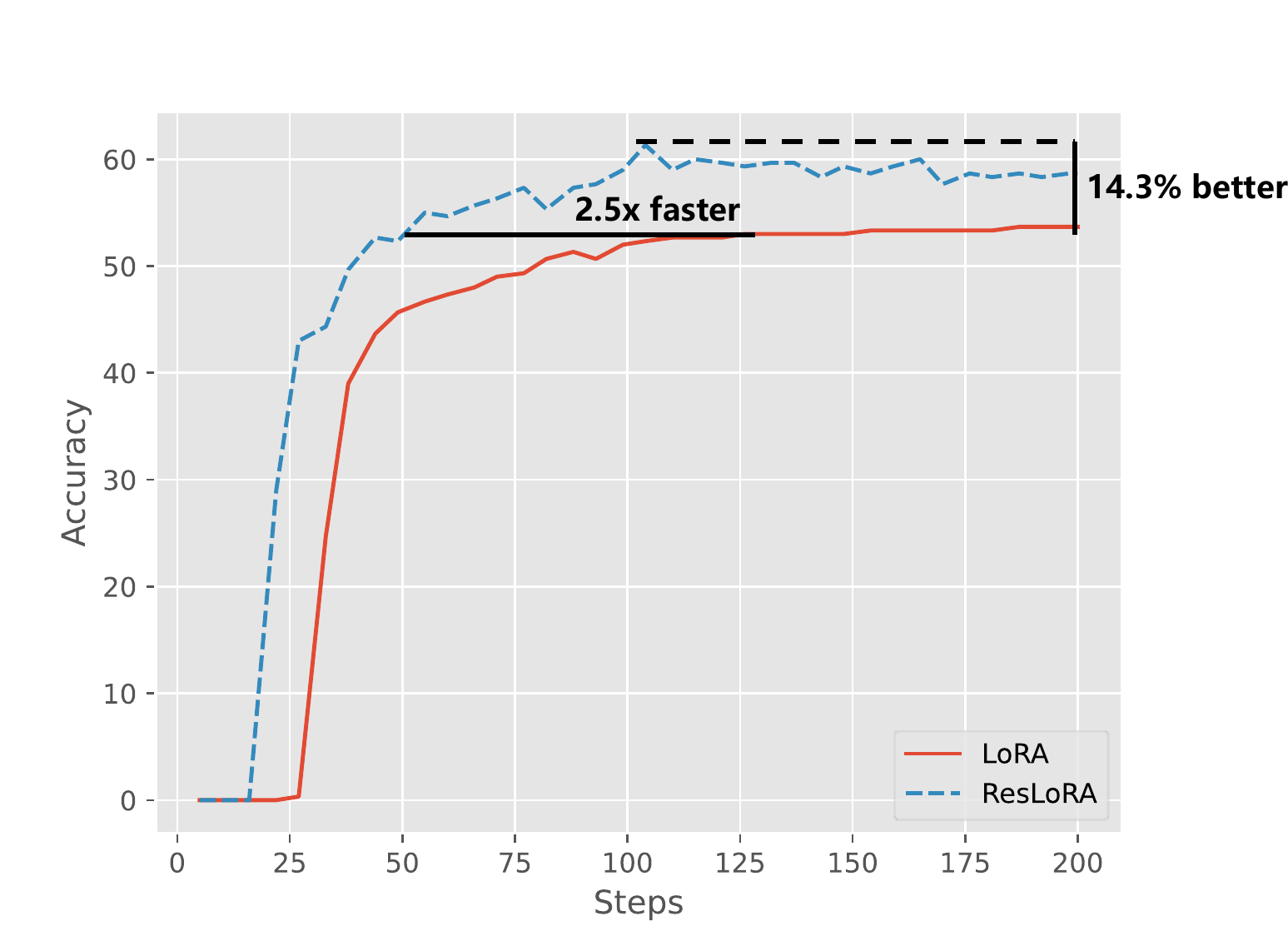}
  \caption{\label{fig:intro}
  An illustration of ResLoRA method on accuracy for SVAMP\citep{svamp}. ResLoRA achieves a 2.5x faster convergence speed and improves performance by 14.3\%.}
\end{figure}

As a prominent method, ResNet\citep{resnet, resnet2} has proven to be widely efficient, and is also used in Transformer models\citep{attention}, between different encoder and decoder blocks. Parallel to linears in these blocks, LoRA blocks can also benefit from the original shortcut design. However, unlike linears, LoRA blocks are more fine-grained. One LoRA block only corresponds to one linear, so the original shortcut is not perfectly suitable for LoRA blocks. Let's use encoders of Transformer as an example. If we add LoRA blocks in query, key, or value linears, the previous gradient must go through the output linear and Softmax function when calculating backward, which may cause gradient vanishing or explosion.

In this paper, we present ResLoRA, a new framework that merges the shortcut of ResNet into LoRA blocks. To validate the efficiency of different residual structures, we propose three residual structures and conduct experiments on different models and tasks. However, the shortcut cannot be directly merged into the original network due to its non-plain structure, which undermines the advantage of LoRA. Considering this, we discuss different merging approaches to convert ResLoRA to the original LoRA blocks, in order to ensure that it can still be merged into the original modules during inference. After merging, our method introduces neither extra parameters nor computational complexity. Compared to LoRA, ResLoRA does not use any additional parameters, but achieves 1\% to 20\% improvement in performance during inference, and lower and faster convergence of loss during training. Moreover, our method can be easily applied to not only the basic LoRA method, but also the other variants of LoRA. Finally, we evaluate the robustness of the results on different experiments. To the best of our knowledge, ResLoRA is the first work that combines the residual path with the LoRA method. The effectiveness of our method is illustrated in Figure \ref{fig:intro}.

Overall, our contribution can be summarized as follows:
\begin{itemize}
    \item We propose ResLoRA, a novel framework that improves LoRA. Compared with the original method, we use residual paths to accelerate the loss reduction process in the training stage, and can usually achieve significant improvements on test datasets.
    \item We investigate different merge approaches for ResLoRA, which can convert it to LoRA blocks, and finally merge them into the frozen original linear, without adding any cost in the inference stage. Benefits from the merge approaches, ResLoRA can be easily applied to other variants of the LoRA method.
    \item We evaluate the results in different models and tasks, to validate the robustness of our improvement. Furthermore, we analyze the reasons why our method can obtain performance gains.
\end{itemize}

\section{Related Works}

\noindent \textbf{Parameter-efficient fine-tuning (PEFT)} Research on PEFT can be divided into three types. One line of research \citep{ptuning, ptuning2} is to add some special trainable vectors attached to the input sequence, which will increase the length of input and have a gap in results compared to full-finetune. Another line of research is to add serialized modules in original modules both in training and inference stage, called Adapter\citep{adapter2, adapter, adapter}. In contrast, LoRA method \citep{lora} adds new low-rank matrices parallel to original linear layer in training stage and merges them into the original model during inference, so there is no extra cost when inferring.


\noindent \textbf{Low-rank training method (LoRA)} Recent studies on LoRA aim to achieve lower cost and better performance. Some researchers explore more flexible and appropriate ranks, such as DyLoRA\citep{dylora}, ReLoRA\citep{relora}, LoHA\citep{loha} and LoKr\citep{lokr}. AdaLoRA\citep{adalora} design a method to dynamically allocate the rank of LoRA blocks in different layers based on their importance, which can reduce the unimportant rank of LoRA blocks. 
Other works focus on the combination of LoRA and other approaches, such as AdaMix\citep{adamix} and QLoRA\citep{qlora}. Besides, LoRAHub\citep{lorahub} and LoRAMoE\citep{loramoe} focus on how to merge multiple LoRA blocks that are fine-tuned on different tasks respectively. 
Despite this, no one has focused on the potential barrier of gradient propagation in LoRA.


\noindent \textbf{Residual Network(ResNet)} \citet{resnet} and previous works\citep{highway} first introduced the residual network. This work solves the gradient vanishing or explosion and improves numerical stability during gradient updating. Considering that the extra shortcut path requires extra computational cost, some works\citep{repvgg} attempt to remove extra paths in the inference stage. Inspired by them. we first extend the main idea of ResNet to LoRA to achieve a faster and more stable training stage, and then design merging approaches to preserve the plain structure of LoRA, so as to maintain the advantages of both LoRA and ResNet at the same time.


\section{Method}

In this section, we introduce our framework, which mainly consists of two parts: (1) ResLoRA blocks, which add various residual paths in LoRA blocks, mainly used in training stage; (2) merging approaches, which remove the residual paths to convert ResLoRA blocks to LoRA blocks, mainly used in inference stage.

\subsection{LoRA Blocks}

We start by revisiting the LoRA method. For an original matrix of the linear layer from a pre-trained model $W_n \in \mathbb{R}^{p \times q}$, where $p$ and $q$ denote the dimensions of output and input, the original equation can be written as $h_n = W_nx_n$, where $x$ denotes the input vector, $h$ denotes the output hidden vector, and $n$ denotes the index of the layer. We define a LoRA block as an additional block parallel to an original matrix. A LoRA block contains two new matrices, down-projection $A \in \mathbb{R}^{r \times p} $ and up-projection $B \in \mathbb{R}^{q \times r}$, which aim to decompose the high-rank matrix into the low-rank matrix. During training, the $W$ is frozen and only the weights of $A$ and $B$ are updated; and during inference, the additional parameters are merged into the original parameters by $W_n+B_nA_n$, to ensure that no latency is introduced. Therefore, the LoRA method can be expressed as Equation \ref{eq:lora}:
\begin{equation} \label{eq:lora}
    h_n = W_nx_n + B_nA_nx_n
\end{equation}
Figure \ref{fig:lora} illustrates the structure of LoRA blocks. Because $r \ll \min(p, q)$, the number of trainable parameters is significantly lower than full fine-tuning.

\subsection{ResLoRA Blocks}
\begin{figure*}[htbp]
  \centering
  \subfloat[LoRA Structure]
  {\includegraphics[scale=.3]{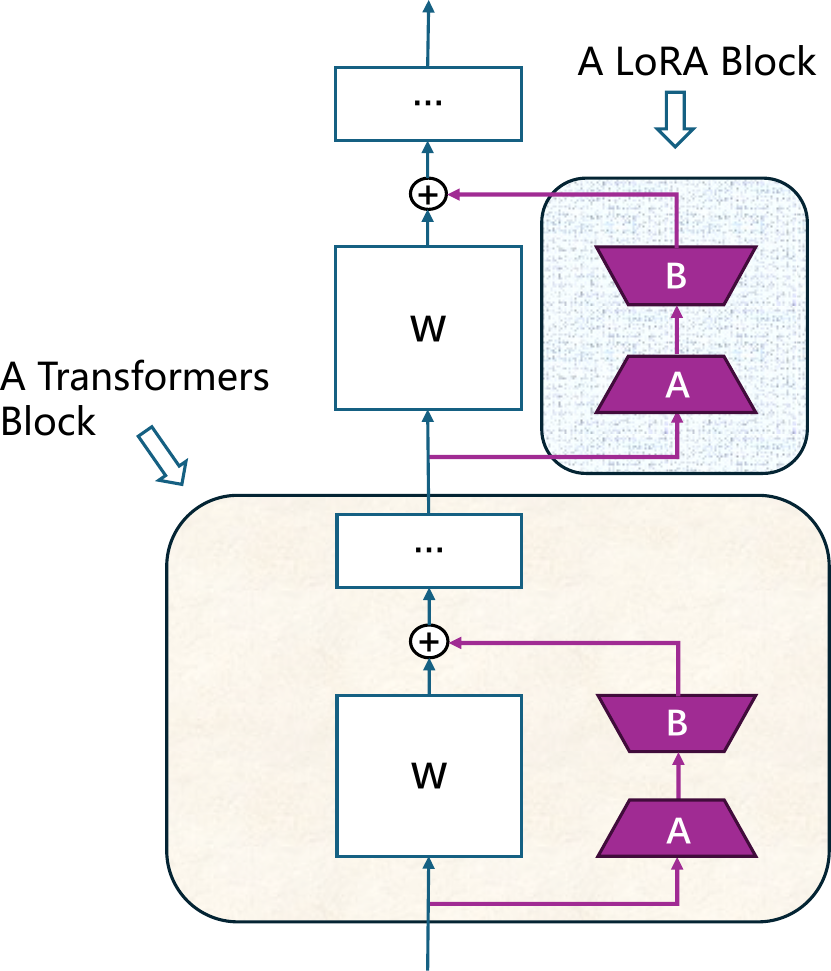}\label{fig:lora}}
  \hspace{.18in} 
  \subfloat[Input-shortcut Structure of ResLoRA\raggedright]
  {\includegraphics[scale=.3]{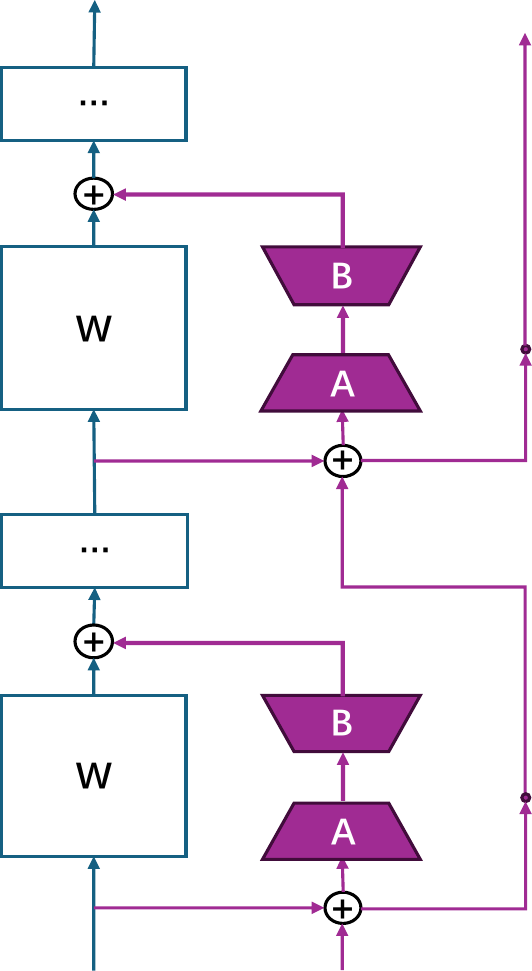}\label{fig:res-1}}
  \hspace{.18in} 
  \subfloat[Block-shortcut Structure of ResLoRA\raggedright]
  {\includegraphics[scale=.3]{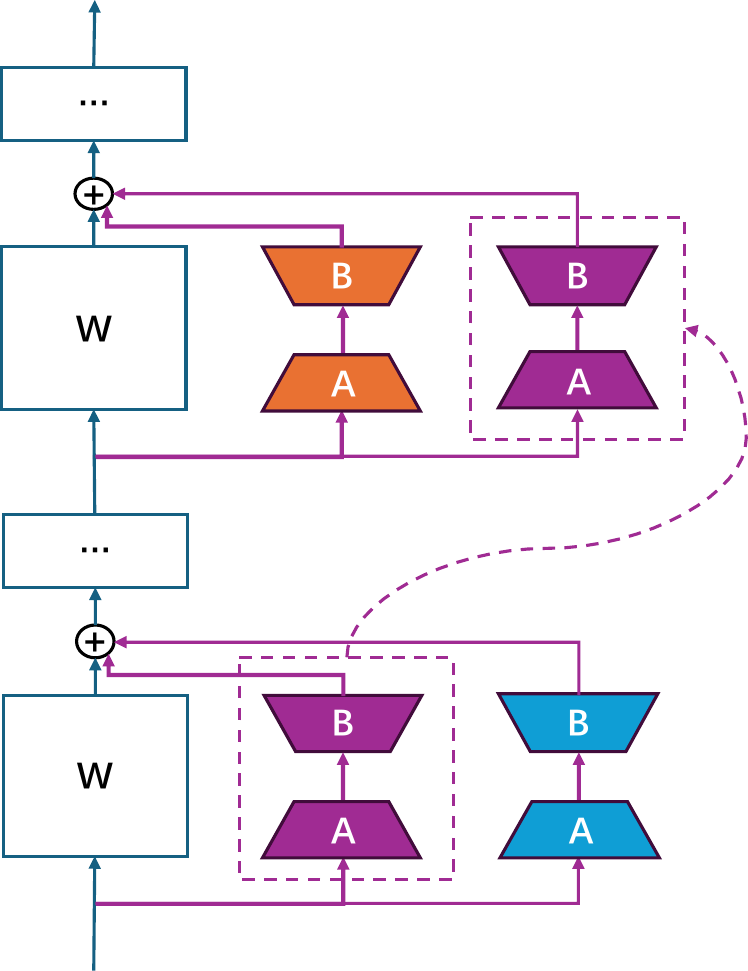}\label{fig:res-2}}
  \hspace{.18in} 
  \subfloat[Middle-shortcut Structure of ResLoRA\raggedright]
  {\includegraphics[scale=.3]{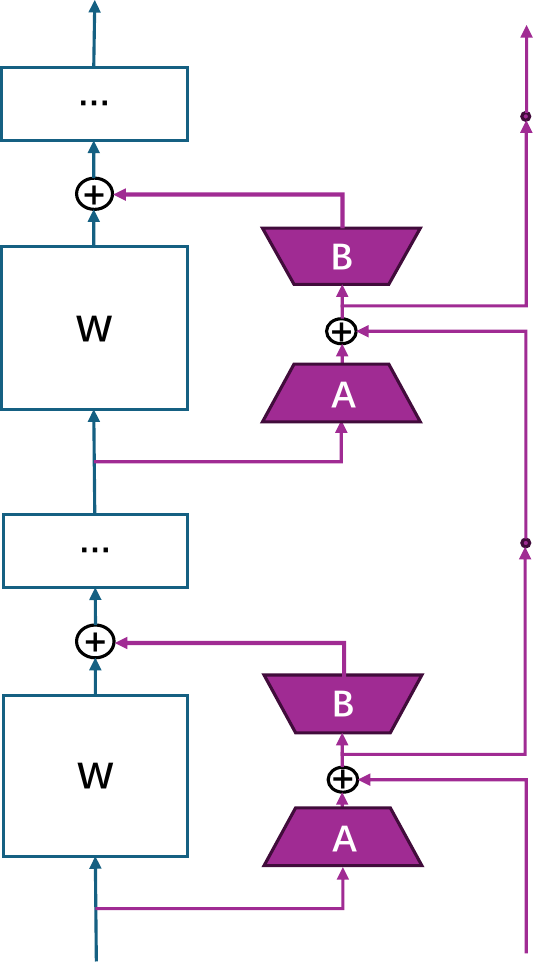}\label{fig:res-3}}
  \caption{\label{fig:reslora}
  Structures of LoRA and ResLoRA}
\end{figure*}

Inspired by ResNet, we introduce residual paths in our method. Considering the different impacts of different structures and mechanisms, we design and implement three types of blocks, named input-shortcut (is), block-shortcut (bs), and middle-shortcut (ms), respectively. Figure \ref{fig:reslora} shows the specific structures of each type.

\noindent \textbf{Input-shortcut structure} means we directly use a shortcut between the input vectors of different LoRA blocks. Specifically, we add the input vector of the previous LoRA block to that of the current LoRA block. This implementation is inspired by the original ResNet, which adds the previous input vectors to current input vectors. Unlike the forward path of ResNet, we don't simply add the input of the LoRA block to the output, because this identity will not only affect the forward path of LoRA blocks, but also the forward path of the original linear layer. As a result, this simple design will create too large values of loss in the forward step to calculate the gradients, thus failing to train LoRA blocks. To avoid this, we only use a shortcut between LoRA blocks. Therefore, the output of a linear layer with the input-shortcut type of ResLoRA can be expressed as:
\begin{equation}
    h_n = W_nx_n + B_nA_n(x_n+x_{n-1})
\end{equation}
where $n \in [1, L]$, and $L$ is the number of layers in original model. If $n=0$, we set $x_{n-1}=x_n$ to maintain the same order of magnitude between the first layer and further layers. The overall structure of this type can be seen in Figure \ref{fig:res-1}.

\noindent \textbf{Block-shortcut structure} means we add shortcuts not to the input vectors, but to the weights of LoRA blocks. Although the input-shortcut structure implements the idea of residual paths, the extra forward path makes it impossible to convert to the original LoRA blocks directly, and merging approaches are required, which may incur performance losses. To obtain the advantages of both the LoRA method and the residual network, we design the block-shortcut structure, which is similar to the DenseNet structure\citep{densenet}. For the input vectors in the current layer, we use the current LoRA block and several previous LoRA blocks simultaneously to participate in the calculation. This forward path does not add any extra forward path for the input, but allows direct transfer of gradients to skip the middle layers, which can also reduce possible obstacles in backward calculation. The output of this structure can be expressed as follows:
\begin{equation}
    h_n = W_nx_n + (\begin{matrix} \sum_{k=0}^m B_{n-k}A_{n-k} \end{matrix}) x_n
\end{equation}
where $m \in [1, L]$ denotes the number of previous LoRA blocks to use. In the specific implementation, we set a hyper-parameter $pre\_num$ to control $m$. Besides, for each layer we set $m=\min(m, n-1)$ to avoid out-of-index errors. Different values of $m$ yield different results, and the details are presented in Section \ref{sec:ablation}. The overall structure of this type is illustrated in Figure \ref{fig:res-2}.

\noindent \textbf{Middle-shortcut structure} means we add shortcuts to the intermediate results of LoRA blocks. LoRA blocks contain two matrices, $A$ and $B$, and $A$ is always closer to the input vectors than $B$. Because of the pre-existing shortcuts between Transformer blocks, $A$ matrices are more likely to benefit from these shortcuts, and are less likely to encounter problems with gradient propagation. Hence we try a new structure so that $B$ matrices can also benefit from the shortcuts. For each layer, the middle-shortcut structure ignores the modification of $A$ matrices, and focuses on the shortcuts of $B$ matrices: we add the previous output vectors of $A$ matrices to the current output vector of $A$ matrix, and then the sum is transferred to the current $B$ matrix. In summary, the process can be expressed as:
\begin{equation}
    h_n = W_nx_n + B_n(\begin{matrix} \sum_{k=0}^m A_{n-k}x_{n-k} \end{matrix})
\end{equation}
Similar to the block-shortcut structure, this structure also uses a hyper-parameter 
$pre\_num$ to control the value of $m$. The overall structure of this type can be seen in Figure \ref{fig:res-3}.

Different types of structures attempt to add shortcuts to different positions, including input vectors, weights of LoRA blocks, and middle vectors, to find the optimal one. The main idea of these structures is the same as ResNet: use shortcuts to reduce the number of modules passed through in backward steps. In the following sections, we use ResLoRA$_{is}$ for the input-shortcut type of ResLoRA, ResLoRA$_{bs}$ for the block-shortcut type, and ResLoRA$_{ms}$ for the middle-shortcut type.

\subsection{Merging Approaches} \label{sec:merge}

While the additional shortcut brings benefits in ResLoRA, there are several issues. One of the most important issues is that a no-plain structure was created. The original LoRA blocks, which we call plain structure, can directly merge into the linear layer, because LoRA doesn't require an extra forward path independent of the original layer. In other words, the forward path of LoRA blocks is similar to the original linear layers. However, ResLoRA uses an additional shortcut between ResLoRA blocks of different layers, which is not the same as the original forward path. Therefore, we need to design merging approaches to convert ResLoRA blocks to LoRA blocks.

How can we convert it? For the block-shortcut structure, the current ResLoRA blocks only require the weights of the previous ResLoRA blocks, and there is no extra forward path, so we can easily merge ResLoRA as follows:
\begin{equation}
    W_n^* = W_n + \begin{matrix} \sum_{k=0}^m A_{n-k}B_{n-k} \end{matrix}
\end{equation}
where $W_n^*$ denotes the weight of the linear layer after merging during inference. However, for the other two structures, direct merging is impossible, because the current ResLoRA blocks require input vectors from previous layers, which create an extra forward path and are different from the original linear layers. Suppose that we can express $x_{n-1}=\alpha x_n$, where $\alpha$ denotes a scaling factor. The previous input vectors can be easily converted to the current input vector, and the ResLoRA blocks can be converted to the LoRA blocks. For different $x_{n-1}$, there is a different $\alpha$ that satisfies $x_{n-1}=\alpha x_n$, so we cannot obtain a precise $\alpha$. Our goal is to find an $\alpha^*$ that satisfies $x_{n-1} \approx \alpha^* x_n$. For example, for the block-shortcut structure, we can derive the following formula:
\begin{equation}
\begin{split}
    h_n &= W_nx_n + B_nA_n(x_n+x_{n-1}) \\
    &\approx W_nx_n + B_nA_n(x_n+\alpha^*x_n) \\
    &= W_nx_n + (1+\alpha^*)B_nA_nx_n
\end{split}
\end{equation}
Therefore, new weights of the current linear layer can be expressed as:
\begin{equation}
    W_n^* = W_n+(1+\alpha^*)B_nA_n
\end{equation}

The precision of $\alpha^*$ is crucial for model inference because this factor directly determines whether the model merging is correct. Since the Frobenius norm, one of the most common matrix norms, can generally measure the size of a matrix\citep{mathbook}, we design two approaches to estimate the value of $\alpha^*$ using the Frobenius norm.

\noindent \textbf{Merge Based on Input.} One approach is to directly calculate $\alpha^*$ based on $x_n$ and $x_{n-1}$. In the training stage, for each layer we use a sliding window to collect the most recent input vectors $x_n$. After that in the inference stage, we calculate the Frobenius norms of all input vectors, and get the average of the Frobenius norms in each sliding window, where $f_n$ denotes the average of Frobenius norms in the n-th layer. We think of this number as representing the size of input vectors, and this mathematical relationship can be expressed as:
\begin{equation}
    \frac{x_n}{f_n} \approx \frac{x_{n-1}}{f_{n-1}}
\end{equation}
Based on this, we can get $\alpha^*$ to be:
\begin{equation}
    \alpha^* = \frac{f_{n-1}}{f_n}
\end{equation}

\noindent \textbf{Merge Based on Weights of ResLoRA Blocks.} Another approach is to get $\alpha^*$ based on the weights of previous ResLoRA blocks rather than input vectors. If we approximate the difference of functions between $x_{n-1}$ and $h_{n-2}$, we can assume that $h_{n-2}$ is the representation of $x_{n-1}$ on the orders of magnitude. Furthermore, if we only focus on the effect of weights and neglect the effect of input vectors, we can get the Frobenius norms of previous weights to represent the $x_{n-1}$ on the orders of magnitude, where $f_{n-2}^*$ denotes the Frobenius norms of weights of $W_{n-1}^*$, which corresponds to the linear layer after merging. Therefore, this relationship between $x_n$ and $x_{n-1}$ can be expressed as:
\begin{equation}
    \frac{x_n}{x_{n-1}} \approx \frac{f_{n-1}^*}{f_{n-2}^*}
\end{equation}
Based on this, we can get $\alpha^*$ to be:
\begin{equation}
    \alpha^* = \frac{f_{n-2}^*}{f_{n-1}^*}
\end{equation}

For ResLoRA$_{ms}$, we simply modify those merging approaches to adapt to the new structure.
For merge based on input, we compute $\alpha^*$ by:
\begin{equation}
    \alpha^* = \begin{matrix} \sum_{k=1}^m \alpha_{n-k}^* \end{matrix}
\end{equation}
where $\alpha_{n-k}^* = f_{n-k} / f_n$, and $f_n$ means the Frobenius norm of weights of $A_nx_n$.
For merge based on weights of previous ResLoRA blocks, we calculate $\alpha$ in the same way and then merge $A$ as:
\begin{equation}
    A^*_n = A_n + \begin{matrix} \sum_{k=1}^m \alpha_{n-k}^*A_{n-k} \end{matrix}
\end{equation}
where $A^*_n$ denotes $A$ after merging, and $\alpha_{n-1}$ denotes $f_{n-1}/f_n$. 

By using these approaches, the ResLoRA blocks can be converted to LoRA blocks, which means no latency will appear in the inference stage. In what follows, we use merge$_{bi}$ for the merge based on input, and merge$_{bw}$ for the merge based on weights of ResLoRA blocks.

\subsection{Mathematics Analyse} \label{sec:math}
Although our method can intuitively solve the potential problem in the backward pass and accelerate the model training process, can we prove it mathematically? For simplicity, we choose the input-shortcut structure as an example. For a specific input $x$, if we want to update the weight of $B_{n-2}$ where $n$ denotes the index of the layer, the gradient can be computed as follows:
\begin{equation}
    \frac{\partial \mathcal{L}}{\partial B_{n-2}} = \frac{\partial \mathcal{L}}{\partial h_n} \frac{\partial h_n}{\partial x_{n-1}} \frac{\partial x_{n-1}}{\partial B_{n-2}}
\end{equation}
where $\mathcal{L}$ is the value of the loss function for this input. For LoRA blocks, the sub-equation is:
\begin{equation} \label{eq:partial-1}
\begin{split}
    \frac{\partial h_n}{\partial x_{n-1}} &= \frac{\partial h_n}{\partial x_n} \frac{\partial x_n}{\partial x_{n-1}} \\
    &= \frac{\partial(W_nx_n+B_nA_nx_n)}{\partial x_n} \frac{\partial x_n}{\partial x_{n-1}}
\end{split}
\end{equation}

However, for the input-shortcut structure of ResLoRA, the sub-equation is:
\begin{equation} \label{eq:partial-2}
\begin{split}
    \frac{\partial h_n}{\partial x_{n-1}} &= \frac{\partial h_n}{\partial x_n} \frac{\partial x_n}{\partial x_{n-1}} \\
    &= \frac{\partial(W_nx_n+B_nA_nx_n)}{\partial x_n} \frac{\partial x_n}{\partial x_{n-1}} \\
    &+ \frac{\partial(B_nA_nx_{n-1})}{\partial x_{n-1}}
\end{split}
\end{equation}

Comparing Equation \ref{eq:partial-2} with Equation \ref{eq:partial-1}, there is no factor before the extra term in Equation \ref{eq:partial-2}, so the gradient can avoid potential vanishing or explosion problems that appear in the factor $\frac{\partial x_n}{\partial x_{n-1}}$. Therefore, the training stage can benefit from ResLoRA blocks.

\section{Experiments}
\subsection{Experimental Setup}
To evaluate the effectiveness of ResLoRA, we conduct experiments on a wide range of models and tasks, including natural language generation (NLG), natural language understanding (NLU), and image generation. We present the details of the tasks in the following subsections respectively.

We compare our method with LoRA\citep{lora}, AdaLoRA\citep{adalora}, LoHA\citep{loha} and LoKr\citep{lokr}, which we detailedly describe in Section \ref{sec:baseline}. All details of experiments can be found in Section \ref{sec:exp-details}.

\subsection{Natural Language Generating}

\noindent \textbf{Models and Datasets.} Considering that LoRA has been mainly used in LLMs recently, we choose LLaMA2-7B\citep{llama2}, a popular open-source LLM, as the NLG model. We conduct experiments on five tasks, including mathematical and commonsense reasoning, which are the primary benchmarks to evaluate the general ability of LLMs. A summary of the datasets is presented in Section \ref{sec:datasets}.


\noindent \textbf{Main Results.} We compare our method with various baseline methods. Table \ref{tab:nlg-result} shows the results of different tasks and methods. LoRA$_{r=16}$ shows significant improvement over LoRA$_{r=4}$ in all tasks, which means that a higher value of rank is effective on these tasks. Among the baseline methods, the original LoRA method performs the best, which may be because the variant methods of LoRA introduce several hyper-parameters. Compared with LoRA, ResLoRA$_{is}$ and ResLoRA$_{bs}$ perform better on almost all tasks. For example, all three types of ResLoRA achieve higher accuracy on HellaSwag, which is 10.98\%, 20.77\%, and 36.85\% higher than LoRA respectively.

\begin{table}
\centering
\resizebox{\linewidth}{!}{
    \begin{tabular}{l|ccccc}
    \toprule
    \textbf{Method} & \textbf{GSM8K} & \textbf{SVAMP} & \textbf{MQA} & \textbf{MMQA} & \textbf{HS} \\
    \midrule
    LoRA$_{r=16}$ & 32.90 & 58.00 & 30.32 & 47.76 & 57.64 \\
    LoRA$_{r=4}$ & 30.93 & 53.33 & 25.43 & 42.23 & 51.36 \\
    AdaLoRA & 15.31 & X & 17.62 & 25.17 & X \\
    LoHA & 19.79 & X & 19.46 & 30.17 & 50.44 \\
    LoKr & 18.5 & X & 18.69 & 21.76 & X \\
    \midrule
    ResLoRA$_{is}$ & 30.33 & 58.33 & \textbf{26.00} & 43.37 & 62.34 \\
    ResLoRA$_{bs}$ & \textbf{31.31} & \textbf{58.67} & 24.86 & \textbf{43.90} & 72.13 \\
    ResLoRA$_{ms}$ & 24.64 & 49.00 & 23.42 & 33.13 & \textbf{88.21} \\
    \bottomrule
    \end{tabular}
    }
\caption{\label{tab:nlg-result} 
Main results of fine-tuning LLaMA2 on NLG tasks. The “X” in the table indicates that the model after fine-tuning does not produce the output according to the instructions, which leads to the incorrect processing of the output. MQA is MathQA, MMQA is MetaMathQA and HS is HellaSwag
}
\end{table}

\subsection{Natural Language Understanding}
\noindent \textbf{Models and Datasets.} We evaluate the proposed methods with RoBERTa-large\citep{roberta} on the General Language Understanding Evaluation (GLUE, \citet{glue}) benchmark, where the model and datasets are the same as \citet{lora}. RoBERTa is a competitive pre-trained model improved from BERT\citep{bert}. GLUE contains different types of tasks, and is widely used to evaluate the NLU ability of models.


\noindent \textbf{Main Results.} Table \ref{tab:nlu-result} shows the results of NLU tasks. The conclusions are similar to those of NLG tasks. LoRA$_{r=16}$ shows significant improvement over LoRA$_{r=4}$ in all tasks, and LoRA$_{r=4}$ shows the best results among all baseline methods. Compared with the baseline methods, our method shows significant improvement on almost all tasks. For all tasks except WNLI, our method demonstrates different degrees of performance enhancement. These experiments verify the general applicability of our method to the NLU tasks.

\begin{table*}
\centering
\resizebox{0.61\linewidth}{!}{
    \begin{tabular}{l|ccccccccc}
    \toprule
    \textbf{Method} & \textbf{MNLI} & \textbf{SST-2} & \textbf{MRPC} & \textbf{CoLA} & \textbf{QNLI} & \textbf{QQP} & \textbf{RTE} & \textbf{STS-B}\\
    \midrule
    LoRA$_{r=16}$ & 90.59 & 95.99 & 91.96 & 67.49 & 94.20 & 91.68 & 83.03 & 92.14\\
    LoRA$_{r=4}$ & 90.26 & 95.30 & 91.13 & 65.29 & 94.23 & 91.01 & 79.06 & 91.67\\
    AdaLoRA & 89.11 & 95.30 & 81.22 & 57.01 & 94.29 & 89.77 & 52.71 & 89.76\\
    LoHA & 89.63 & 95.76 & 88.85 & 60.57 & 94.03 & 89.93 & 55.23 & 90.21\\
    LoKr & 87.08 & 94.50 & 81.41 & 55.22 & 92.44 & 88.30 & 52.71 & 83.35\\
    \midrule
    ResLoRA$_{is}$ & 89.97 & 95.64 & \textbf{92.39} & 65.54 & 94.34 & 90.91 & \textbf{83.03} & \textbf{91.97}\\
    ResLoRA$_{bs}$ & \textbf{90.40} & \textbf{96.22} & 91.31 & 65.44 & \textbf{94.48} & \textbf{91.27} & 82.31 & 91.72\\
    ResLoRA$_{ms}$ & 88.74 & 95.76 & 91.64 & \textbf{65.80} & 94.34 & 87.14 & 81.95 & 91.11 \\
    \bottomrule
    \end{tabular}
    }
\caption{\label{tab:nlu-result}
Main results of fine-tuning RoBERTa-large on NLU tasks. 
}
\end{table*}

\subsection{Text to Image}
\noindent \textbf{Models and Datasets.} To verify the generalization of our method, we also conduct experiments on multi-modal tasks. We select the text-to-image task, which aims to generate the appropriate images based on input texts. The model we use is the popular Stable-Diffusion-v2\citep{sd-2}, one of the most popular image generation models. The dataset we select is \citet{pokemon}, which contains images of a cartoon style. Our goal is to let the model learn this cartoon style, which is greatly different from its original style.

\begin{figure*}[htbp]
  \centering
  \subfloat[Training datasets]
  {\includegraphics[scale=.22]{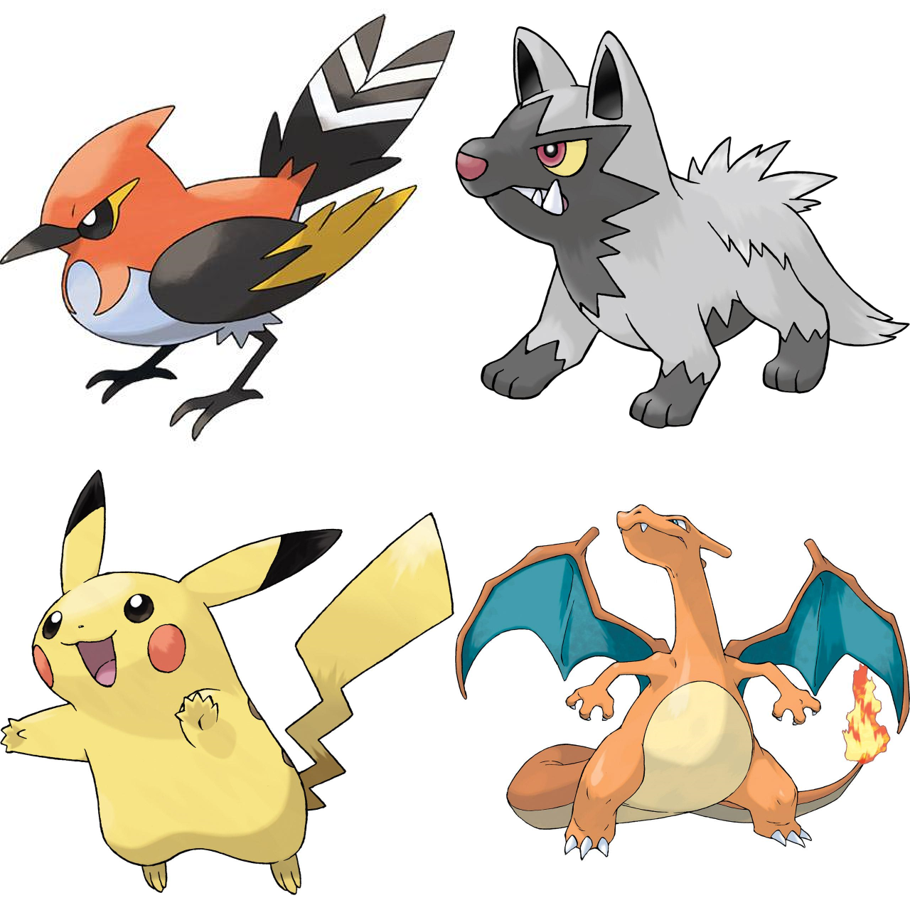}\label{fig:diff_0}}
  \hspace{.01in} 
  \subfloat[Prompt: a gray and black cat with yellow eyes]
  {\includegraphics[scale=.22]{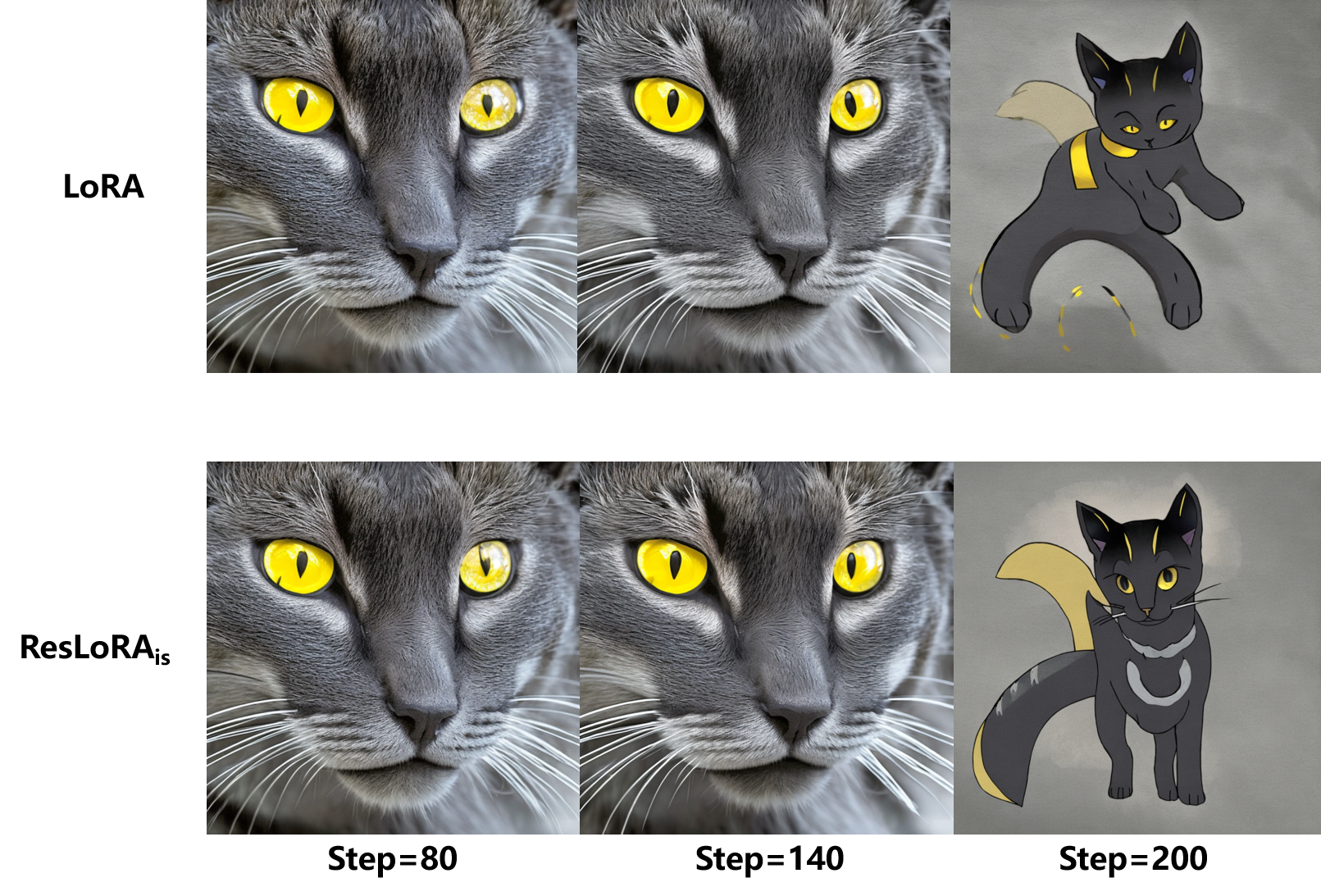}\label{fig:diff_1}}
  \hspace{.01in} 
  \subfloat[Prompt: a small bird with a black and white tail]
  {\includegraphics[scale=.22]{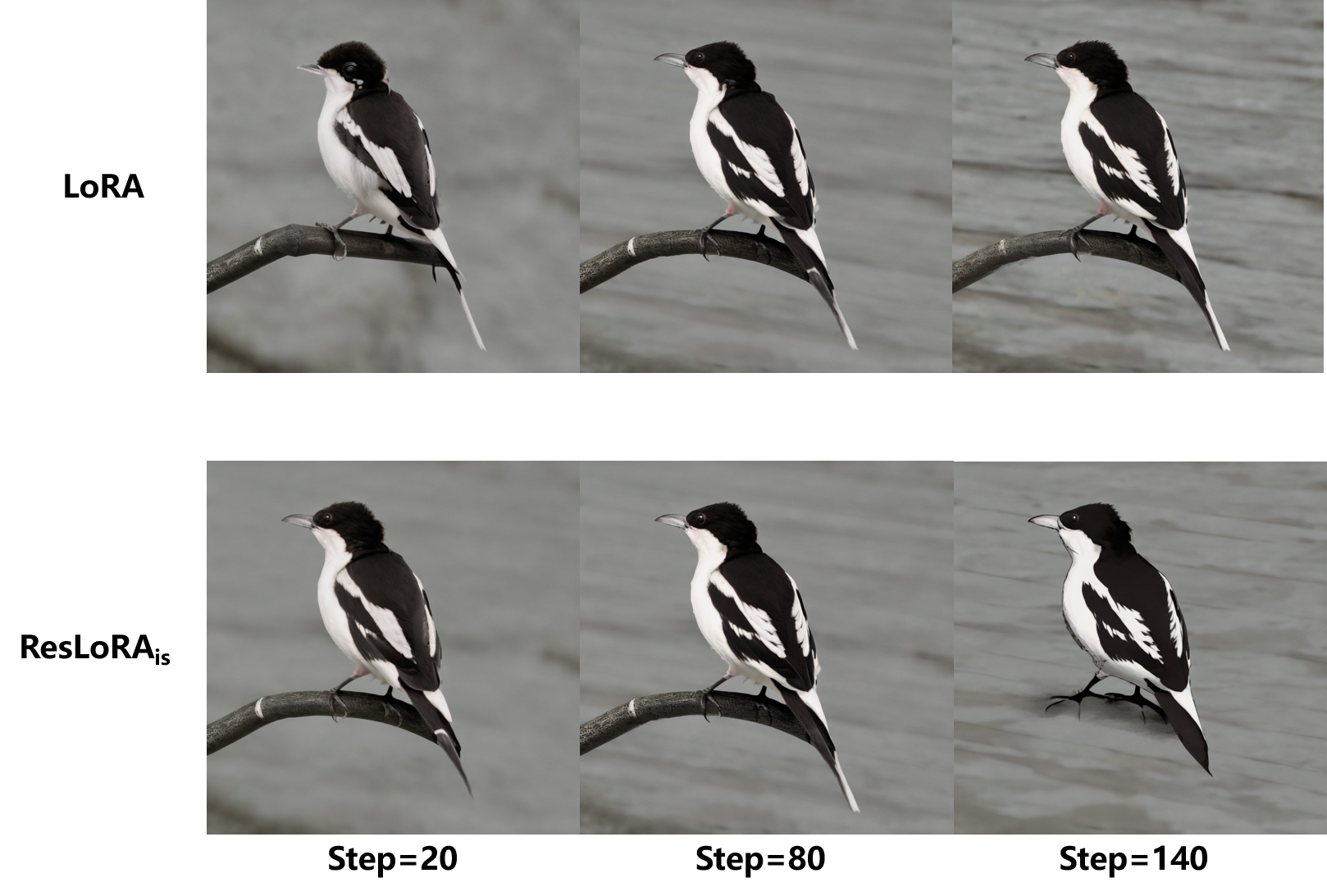}\label{fig:diff_2}}
  \caption{\label{fig:diff}
  Results of text-to-image task. We compare images generated by LoRA and ResLoRA$_{is}$.}
\end{figure*}

\noindent \textbf{Main Results.} Figure \ref{fig:diff} shows the results of text-to-image task. We use two prompts to generate images that are trained by LoRA and ResLoRA$_{is}$ respectively, and save the images from the training process. For both groups of images, ResLoRA$_{is}$ clearly generates more appropriate images in the later step. In Figure \ref{fig:diff_1}, the 200-step result of ResLoRA$_{is}$ is already a vivid cartoon cat with legs, a tail, and clothes, whereas the result of the same step in LoRA is still an incomplete character. In Figure \ref{fig:diff_2}, the 140-step result of ResLoRA$_{is}$ has been converted to the Pokemon style, while the result of LoRA is still a realistic bird, which is not our goal. In short, ResLoRA$_{is}$ shows better results and a faster training process for this task.

\subsection{Ablation studies}
\label{sec:ablation}

\noindent \textbf{Merge Approach.} In Section \ref{sec:merge}, we propose two merge approaches to convert ResLoRA blocks to LoRA blocks, to avoid the extra cost in the inference stage. However, our approaches introduce unavoidable accuracy degradation. Nevertheless, our approaches achieve higher accuracy than LoRA and other variants.

Table \ref{tab:merge} shows the original accuracy before merging, and the results after merging. The huge gap between the results before merging and the results with no merging approach confirms our point: merging approaches are essential for ResLoRA, because the additional residual path is introduced in the training stage, which causes the difference between training and inference. In spite of the different calculations, the results of the two approaches are similar in ResLoRA$_{is}$. Our two merging approaches both have an accuracy degradation of about 1\% compared to the results before merging, and have an accuracy improvement of about 10\% compared to the results with no merging.

However, for ResLoRA$_{ms}$ there is a large gap between the results with and without merging. This may be because of more error accumulation and greater difficulty when merging more previous blocks. Therefore, it is worthwhile to continue to improve merging approaches. For the two modified approaches, merge$_{bw}$ shows worse results than merge$_{bi}$. More detailed results in different datasets can be found in Section \ref{sec:merge-result}.

\begin{table}
\centering
\resizebox{\linewidth}{!}{
    \begin{tabular}{l|cccc}
    \toprule
     \multirow{2}{*}{\textbf{Merge Approach}} & \multicolumn{2}{c}{\textbf{ResLoRA$_{is}$}} & \multicolumn{2}{c}{\textbf{ResLoRA$_{ms}$}}\\
     & \textbf{GSM8K} & \textbf{SVAMP} & \textbf{GSM8K} & \textbf{SVAMP} \\
    \midrule
    LoRA$_{r=4}$ & 30.93 & 53.33 & - & - \\
    \midrule
    w/o merge & 30.48 & 60.00 & 31.01 & 60.33 \\
    merge$_{no}$ & 20.85 & 48.00 & - & - \\
    merge$_{bi}$ & 29.49 & 59.00 & 24.64 & 49.00 \\
    merge$_{bw}$ & 30.33 & 58.33 & 18.35 & 51.33 \\
    \bottomrule
    \end{tabular}
    }
\caption{\label{tab:merge}
The results of the different merging approaches for ResLoRA$_{is}$. Merge$_{no}$ means we do not apply any merging approach and simply use the trained weights to infer as LoRA blocks. Merge$_{bi}$ and merge$_{bw}$ are what we mentioned in Section \ref{sec:merge}. For ResLoRA$_{ms}$, we report the results with $pre\_num=4$.
}
\end{table}

\noindent \textbf{The Number of Previous ResLoRA Blocks.} In ResLoRA$_{bs}$ and ResLoRA$_{ms}$, we intend to use not only one adjacent previous block but also multiple blocks to calculate the result. To verify the effectiveness of different numbers of previous blocks, we compare their results in Table \ref{tab:prenum}. From the table, we observe that the value of $pre\_num$ affects the result significantly. Too large or too small a value is harmful, and the best results occur with a proper value of $pre\_num$. Besides, the results show that it is workable to incorporate more previous blocks because more potential obstacles are skipped when training. The results of ResLoRA$_{ms}$ are similar to those of ResLoRA$_{bs}$.

\begin{table}
\centering
\resizebox{\linewidth}{!}{
    \begin{tabular}{c|cccc}
    \toprule
    \multirow{2}{*}{\textbf{Previous blocks}} & \multicolumn{2}{c}{\textbf{ResLoRA$_{is}$}} & \multicolumn{2}{c}{\textbf{ResLoRA$_{ms}$}}\\
    & \textbf{GSM8K} & \textbf{SVAMP} & \textbf{GSM8K} & \textbf{SVAMP} \\
    \midrule
    0 & 30.93 & 53.33 & - & - \\
    \midrule
    1 & 30.78 & 57.00 & 31.16 & 59.00 \\
    2 & 30.25 & 57.33 & 30.25 & 59.33 \\
    3 & 30.71 & 58.67 & 30.48 & 59.33 \\
    4 & 31.31 & 58.67 & 31.01 & 60.33 \\
    5 & 32.98 & 57.67 & 31.08 & 61.33 \\
    -1 & 30.02 & 56.00 & 30.02 & 58.67 \\
    \bottomrule
    \end{tabular}
    }
\caption{\label{tab:prenum}
The results with different numbers of previous blocks in ResLoRA$_{bs}$. 
$0$ means the ResLoRA degenerates to the original LoRA method. $-1$ means $\infty$, where each ResLoRA block uses all previous blocks. To avoid the impact of the merge approaches, we report the results of ResLoRA$_{ms}$ before merging.
}
\end{table}

\subsection{Analysis}
\label{sec:analyse}

\noindent \textbf{Extra residual paths can accelerate the training stage.} As we showed in Section \ref{sec:math}, extra residual paths can speed up the training stage and achieve better fitness of the models. But is that really true? To verify this conclusion, we collect the loss curves.

\begin{figure}[htbp]
  \centering
  \includegraphics[width=\linewidth]{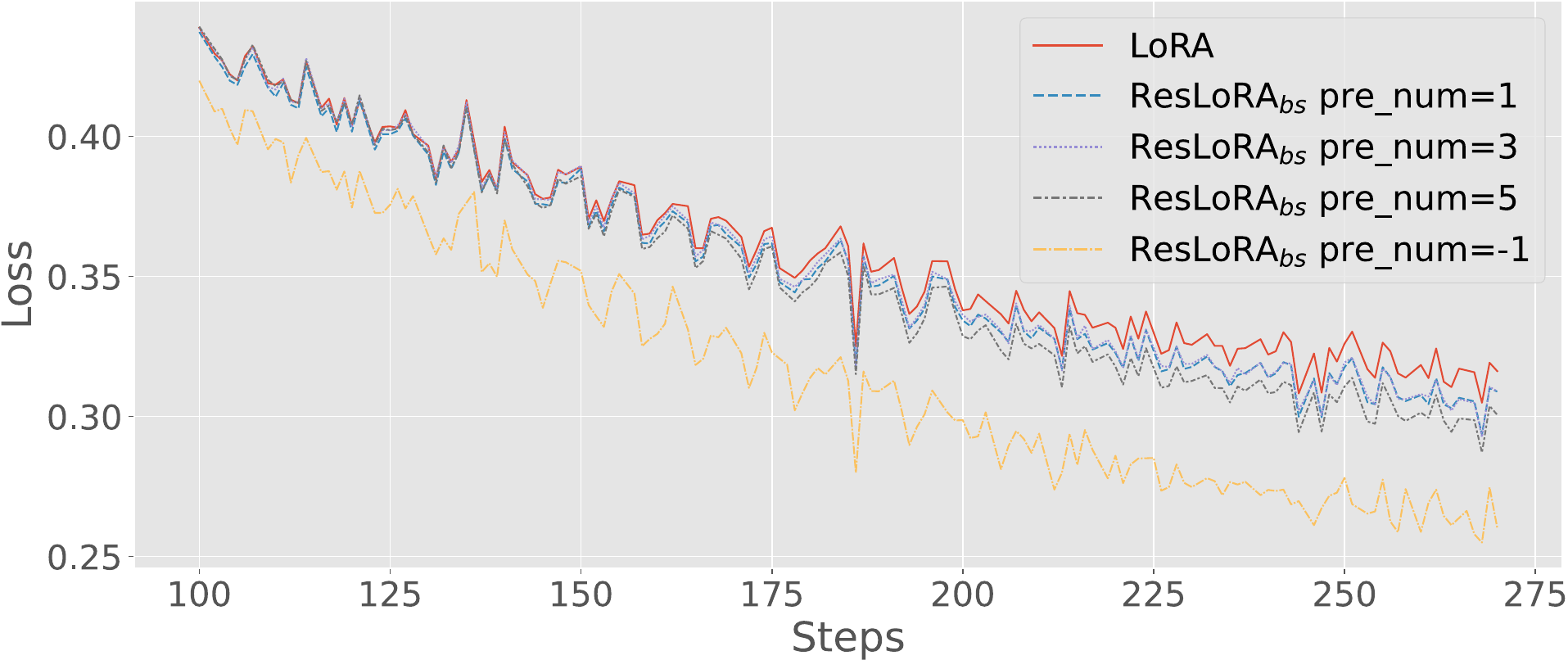}
  \caption{\label{fig:loss}
  Training loss with different $pre\_num$ values on SVAMP. 
  $pre\_num=-1$ means each ResLoRA block uses all previous ResLoRA blocks.}
\end{figure}

Figure \ref{fig:loss} shows the partial process of the loss when training with LoRA and ResLoRA$_{bs}$ with different $pre\_num$, which denotes the number of previous ResLoRA blocks. The original LoRA method can be considered as ResLoRA$_{bs}$ with $pre\_num=0$. As we can see in the figure, the LoRA method has the largest loss and the ResLoRA with $pre\_num=-1$ has the smallest loss. Moreover, as $pre\_num$ increases, the loss decreases faster and finally reaches the lower value. 

\begin{figure}[htbp]
  \centering
  \includegraphics[width=\linewidth]{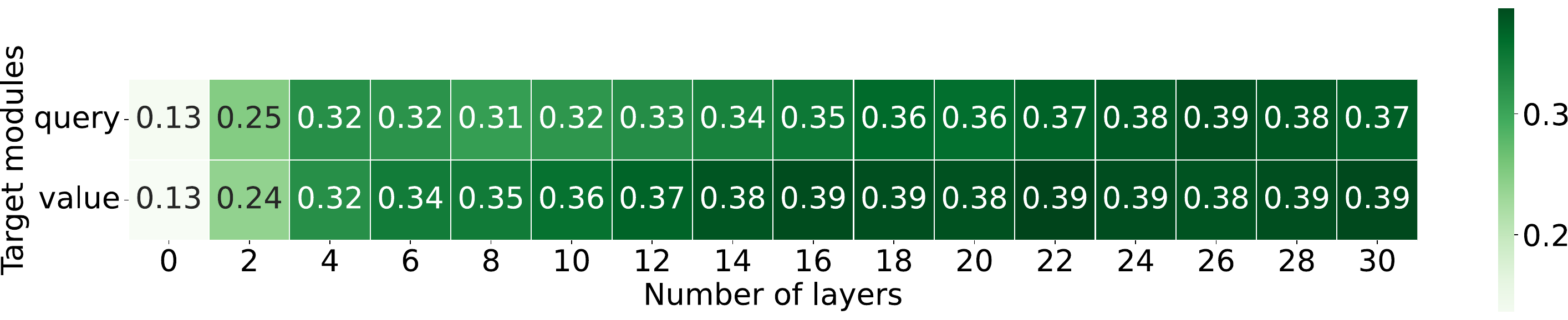}
  \caption{\label{fig:weight}
  Difference of the weights of trained matrices between LoRA and ResLoRA$_{bs}$ blocks. We fine-tune models on SVAMP both for 20 epochs, and observe their difference. The ResLoRA$_{bs}$ blocks have been merged.}
\end{figure}

\noindent \textbf{Extra residual paths produce more complex weights of LoRA matrices.} Despite the mathematical reasoning in Section \ref{sec:math} that proves faster convergence when training, we still don't know what exactly causes the higher accuracy. To explain this result, we save the model weights after sufficiently fine-tuning it, and compare the difference of trained matrices between the two methods. The result is displayed in Figure \ref{fig:weight}.

Considering the high computational complexity of large matrices, we use the Frobenius norm to measure their complexity. In the figure, we subtract the F-norm of each LoRA block from each merged ResLoRA$_{bs}$ block, and show this difference through the heat map. Apparently, ResLoRA blocks contain elements with larger absolute values than LoRA blocks, which implies that the blocks can be trained more adequately using ResLoRA. This may be one of the reasons for the better performance of our method. More detailed results can be found in Section \ref{sec:analyse-extra}.

\section{Conclusion}
We develop a new improved framework ResLoRA for low-rank adaptation (LoRA). ResLoRA adds residual paths during training, and uses merging approaches to remove these paths during inference. Without any extra trainable parameters, ResLoRA can achieve better results in fewer training steps compared to the original LoRA and other baseline methods. We conduct experiments with three types of ResLoRA structures on NLG, NLU, and text-to-image tasks, and the results of almost all tasks verify the effectiveness of our method.

\newpage
\section*{Limitations}
In this section, we discuss some limitations of our method and provide some suggestions for future work.
\begin{enumerate}
    \item Despite adding no extra trainable parameters, the training cost of our method is higher than that of standard LoRA, because ResLoRA needs to use previous blocks when computing in one block. The more previous blocks are used, the higher the cost becomes. Therefore, it is important to balance the training cost and the performance.
    \item During inference, we apply merging approaches to remove the extra residual paths. However, none of the merging approaches we proposed can achieve lossless merging, which compromises the final performance of the model. Designing more efficient approaches is desirable.
    \item Our work is the first to combine residual paths with LoRA. Prior to this, many valuable works have been proposed, such as \citet{adalora, qlora, relora}. Since there is no fundamental conflict between our ResLoRA and other methods, it is feasible to integrate these methods. We leave these investigations for future work.
\end{enumerate}



\section*{Ethics Statement}
This paper proposes a more efficient framework based on LoRA, which is helpful for customizing training models with low resources. We firmly believe that LLMs are not only a potential area of NLP, but also significant for all areas of artificial intelligence. Therefore, it is important to fine-tune a model efficiently. Although fine-tuning is a dual-use technology, we believe in the positive impact of our method.

\bibliography{anthology,custom}
\bibliographystyle{acl_natbib}

\appendix

\section{Baseline}
\label{sec:baseline}

We compare our ResLoRA with the following methods:
\begin{itemize}
    \item LoRA\citep{lora} is what we based on. It's popular for its low memory requirement in the training stage and no latency in the inference stage.
    \item AdaLoRA\citep{adalora} is an important variant of LoRA, which can adjust more ranks dynamically in important layers and modules in the training stage.
    \item LoHA\citep{loha} is another popular variant of LoRA, which uses the Hadamard product of two low-rank matrices to obtain a high-rank matrix, to perform the high-rank updating.
    \item LoKr\citep{lokr} is a method similar to LoHA. The difference is that LoKr replaces the Hadamard product in LoHA with the Kronecker product.
\end{itemize}

\section{Details of NLG datasets}
\label{sec:datasets}
For mathematical tasks, we select four datasets. GSM8K\citep{gsm8k} is a high-quality linguistically diverse dataset of grade school math word problems, which is widely used in open leaderboards. SVAMP\citep{svamp} contains simple math word problems created by applying carefully chosen variations to examples sampled from existing datasets. MathQA\citep{mathqa} is an advanced dataset gathered by using a new representation language to annotate the AQuA-RAT dataset\citep{aqua} with fully specified operational programs. MetaMathQA\citep{mmqa} uses LLMs to rewrite the question from multiple perspectives based on GSM8K\citep{gsm8k} and MATH\citep{mathdataset}.

For commonsense tasks, we select HellaSwag. HellaSwag\citep{hs} is a challenging dataset, which contains questions to select the best endings to complete sentences. It has been considered as one of the most common datasets to judge the reasoning ability of LLMs.

We demonstrate the details of our NLG datasets in Table \ref{tab:overview-datasets}.

\begin{table*}
\centering
    \begin{tabular}{l|cccc}
    \toprule
    \textbf{Datasets} & \textbf{Domain} & \textbf{Scale} & \textbf{Open-ended} & \textbf{Reasoning Process} \\
    \midrule
    GSM8K & Math & 8K & Yes & Yes \\
    SVAMP & Math & 1K & Yes & Yes \\
    MathQA & Math & 30K & No & Yes \\
    MetaMathQA & Math & 395K & Yes & Yes \\
    HellaSwag & Commonsense & 40K & No & No \\
    \bottomrule
    \end{tabular}
\caption{\label{tab:overview-datasets}
Overview of our NLG datasets. Open-ended means whether the answer is given by options, or required to generate directly. Reasoning process means whether there are reasoning processes in labels for model training.
}
\end{table*}

\section{Experiments Details}
\label{sec:exp-details}

We use the public Pytorch\citep{pytorch} and Huggingface Transformers\citep{transformers} libraries to implement the code of all methods, and conduct all the experiments using Tesla V100 GPUs. For all structures of the ResLoRA method, we set the input vectors of previous blocks to be none for the blocks that don't have a previous one, such as blocks in the first layer of the model. The details of different models are presented respectively. For other baseline methods, we implement them via the PEFT\citep{peft} package from Huggingface. We only consider the linear layers as the target modules.

\noindent \textbf{Implementation Details of NLG Experiments.} To reduce the GPU memory usage when training the LLM, we use the ZeRO-2 stage\citep{zero} to offload unnecessary parameters from GPU to CPU via Deepspeed\citep{deepspeed}. For all experiments except $LoRA_{r=16}$, we set the rank $r=4$, the alpha $\alpha=8$, and the target module as query $W_q$ and value $W_v$ following the setup of \citet{lora}. For MetaMathQA, we randomly select 40k samples to train and test in the test dataset of GSM8K. Moreover, to let the LLM understand and follow the format of inputs and outputs, we use simple prompts for each task. 

For all datasets, we report accuracy as the evaluation metric, which is higher the better. We report the results for ResLoRA$_{is}$ with the merge$_{bw}$, for ResLoRA$_{bs}$ with $pre\_num=4$, and for ResLoRA$_{ms}$ with $pre\_num=4$ after merging with the merge$_{bi}$. More details can be found in our code. 

\noindent \textbf{Implementation Details of NLU Experiments.} We use the same baseline methods and report results with the same hyper-parameters in our method. The only difference is that there is no need to use Deepspeed because RoBERTa-large is not an LLM.

We report the F1 score for MRPC, the Matthew's correlation coefficient for CoLA, the Pearson correlation for STS-B, and the accuracy for others. For all metrics, higher is better. We report the results for ResLoRA$_{is}$ with the merge$_{bw}$, for ResLoRA$_{bs}$ with $pre\_num=4$, and for ResLoRA$_{ms}$ with $pre\_num=4$ after merging with the merge$_{bi}$.

\noindent \textbf{Implementation Details of Text-to-image Experiments.} There are three parts in Stable-Diffusion, including the text encoder, the variational auto-encoder\citep{vae} and the U-Net\citep{unet}. We apply our methods to U-Net. Considering the different sizes of matrices in adjacent layers in U-Net, we add LoRA blocks in all of $\{W_q, W_k, W_v, W_o\}$, but only enable residual path in blocks that have the same size as neighboring previous blocks. Moreover, we set the rank $r=16$ and the alpha $\alpha=32$ to produce more prominent results.


\section{Results for Merging Approach}
\label{sec:merge-result}

We show the detailed results of merging experiments. Results of ResLoRA$_{is}$ are in Table \ref{tab:merge-res-1}, and the results of ResLoRA$_{ms}$ are in Table \ref{tab:merge-res-3}. There is no need to apply extra merging approaches for ResLoRA$_{bs}$.

\begin{table*}
\centering
    \begin{tabular}{l|ccccc}
    \toprule
    \textbf{Merge Approach} & \textbf{GSM8K} & \textbf{SVAMP} & \textbf{MathQA} & \textbf{MetaMathQA} & \textbf{HellaSwag} \\    
    \midrule
    LoRA$_{r=4}$ & 30.93 & 53.33 & 25.43 & 42.23 & 51.36 \\
    w\textbackslash o merge & 30.48 & 60.00 & 26.47 & 43.37 & 67.20 \\
    merge$_{bi}$ & 29.49 & 59.00 & 25.83 & 42.91 & 61.48 \\
    merge$_{bw}$ & 30.33 & 58.33 & 26.00 & 43.37 & 62.34 \\
    \bottomrule
    \end{tabular}
\caption{\label{tab:merge-res-1}
The results with the different merge approaches in ResLoRA$_{is}$. For all datasets we report accuracy.
}
\end{table*}

\begin{table*}
\centering
    \begin{tabular}{l|ccccc}
    \toprule
    \textbf{Merge Approach} & \textbf{GSM8K} & \textbf{SVAMP} & \textbf{MathQA} & \textbf{MetaMathQA} & \textbf{HellaSwag} \\    
    \midrule
    LoRA$_{r=4}$ & 30.93 & 53.33 & 25.43 & 42.23 & 51.36 \\
    w\textbackslash o & 31.01 & 61.33 & 29.15 & 44.88 & 86.13 \\
    merge$_{bi}$ & 24.64 & 49.00 & 23.42 & 33.13 & 88.21 \\
    merge$_{bw}$ & 18.35 & 51.33 & 18.29 & 22.59 & 84.07 \\
    \bottomrule
    \end{tabular}
\caption{\label{tab:merge-res-3}
The results with the different merge approaches in ResLoRA$_{ms}$  with $pre\_{num}=4$. For all datasets, we report accuracy.
}
\end{table*}

\section{Results of Analysis}
\label{sec:analyse-extra}

\begin{figure*}[htbp]
  \centering
  \includegraphics[width=\linewidth]{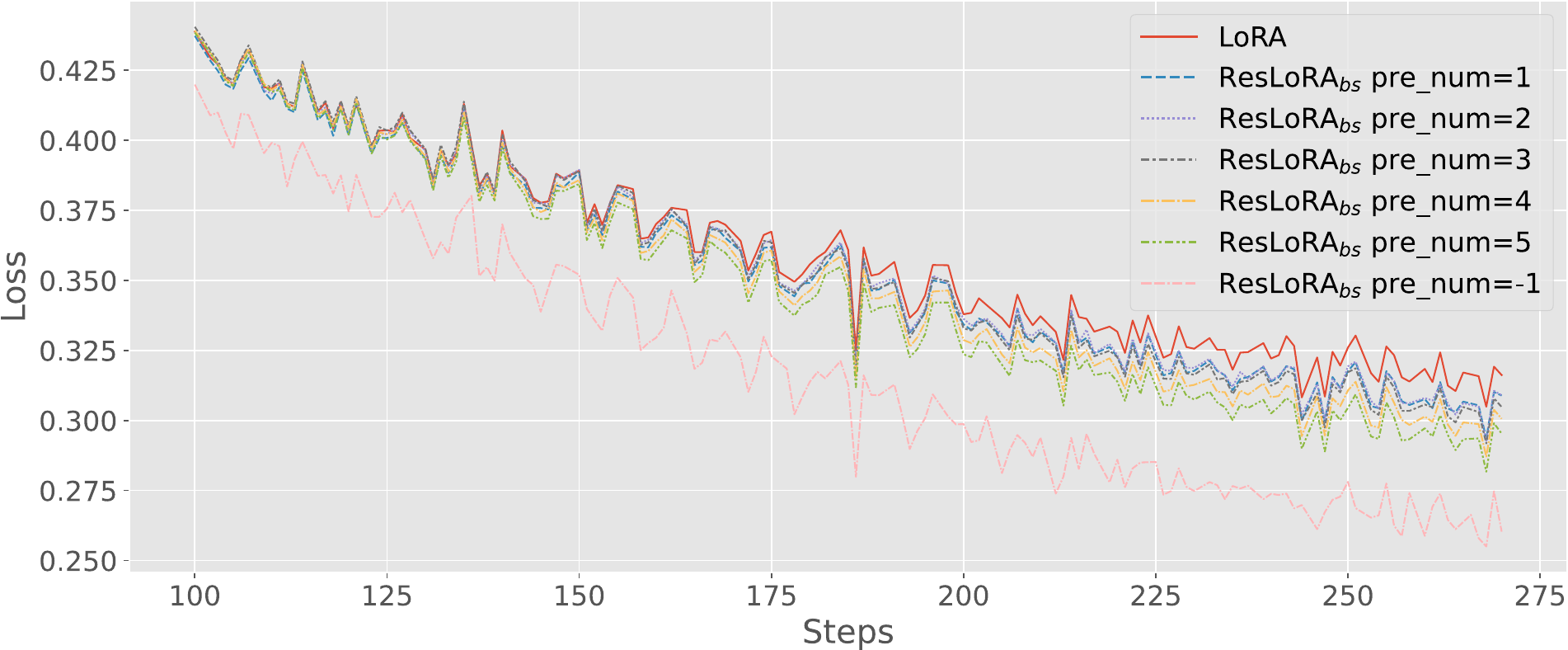}
  \caption{\label{fig:loss_full}
  Training loss with different $pre\_num$ values on SVAMP. $pre\_num$ indicates how many previous blocks will be used, and $pre\_num=-1$ means each ResLoRA block uses all previous ResLoRA blocks.}
\end{figure*}

\begin{figure*}[htbp]
  \centering
  \includegraphics[width=\linewidth]{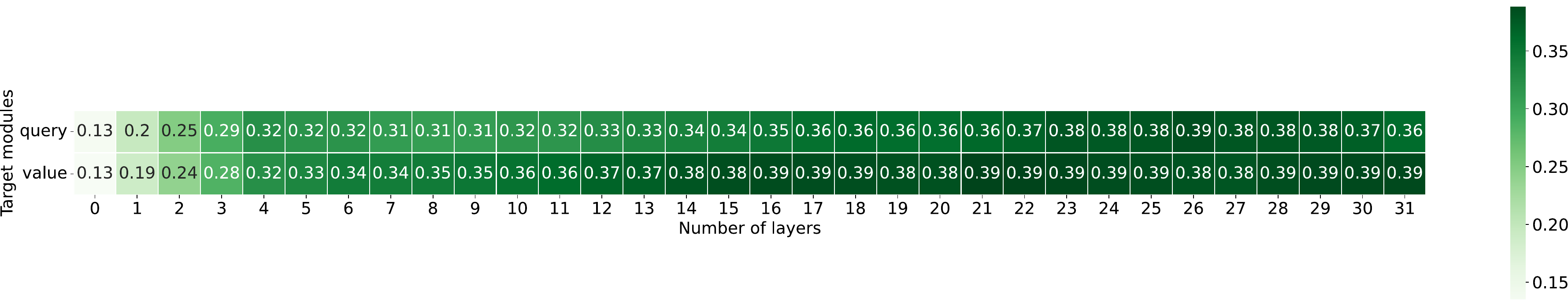}
  \caption{\label{fig:weight_full}
  Difference of the weights of trained matrices between LoRA and ResLoRA$_{bs}$ blocks. We fine-tune models on SVAMP both for 20 epochs, and observe their difference. The ResLoRA$_{bs}$ blocks have been merged.}
\end{figure*}

We show the full version of Figure \ref{fig:loss} in Figure \ref{fig:loss_full}, and the full version of Figure \ref{fig:weight} in Figure \ref{fig:weight_full}.

\end{document}